\documentclass[]{article}
\makeatletter
\renewenvironment{thebibliography}[1]
{\section*{\refname\@mkboth{\refname}{\refname}}%
  \list{\@biblabel{\@arabic\c@enumiv}}%
       {\settowidth\labelwidth{\@biblabel{#1}}%
        \leftmargin\labelwidth
        \advance\leftmargin\labelsep
 \setlength\itemsep{-0.5ex}
 \setlength\baselineskip{8pt}
        \@openbib@code
        \usecounter{enumiv}%
        \let\p@enumiv\@empty
        \renewcommand\theenumiv{\@arabic\c@enumiv}}%
  \sloppy
  \clubpenalty4000
  \@clubpenalty\clubpenalty
  \widowpenalty4000%
  \sfcode`\.\@m}
 {\def\@noitemerr
   {\@latex@warning{Empty `thebibliography' environment}}%
  \endlist}
\makeatother

\newcommand{\lw}[1]{\smash{\lower3.0ex\hbox{#1}}}

\usepackage{amsmath}
\usepackage[dvipdfmx]{graphicx}
\usepackage{bm,amsmath,amssymb}
\usepackage{amsmath,amssymb}
\usepackage{comment}

\renewcommand{\refname}{References}

\usepackage{mathtools}
\usepackage{url}

\begin{document}
\twocolumn[
\noindent

\hspace{1em}
\hfill

\vspace{2mm}

\hrule

\begin{center}
	{\Large \bf Deep Pyramidal Residual Networks with Separated Stochastic Depth}
\end{center}

\hrule
\vspace{10mm}

\begin{center}
	{\large \bf Yoshihiro Yamada, Masakazu Iwamura and Koichi Kise}
\end{center}
\begin{center}
{\large Graduate School of Engineering, Osaka Prefecture University}
\end{center}
\vspace{10mm}

\par]

\begin{center}
	{\Large \bf Abstract}
\end{center}
	{\it On general object recognition, Deep Convolutional Neural Networks (DCNNs) achieve high accuracy.
	In particular, ResNet and its improvements have broken the lowest error rate records.
	In this paper, we propose a method to successfully combine two ResNet improvements, ResDrop and PyramidNet.
	We confirmed that the proposed network outperformed the conventional methods;
	on CIFAR-100, the proposed network achieved an error rate of 16.18\% in contrast to PiramidNet achieving that of 18.29\% and ResNeXt 17.31\%.}

\section{Introduction}
ResNet~\cite{He_CVPR2016} is a well-known deep convolutional neural network (DCNN) because ResNet and its improvements~\cite{arXiv:1603.09382,arXiv:1608.06993,arXiv:1610.02915,arXiv:1611.05431} have broken the lowest error rate records.
It is known that a deeper network can have a higher discriminant ability.
However, realizing it is difficult because of nuisances such as the vanishing gradient problem.
To avoid them, ResNet introduces a processing block called residual block to facilitate learning of a deeper network. \\
ResDrop is a ResNet improvement which further avoids the vanishing gradient problem~\cite{arXiv:1603.09382}.
In DCNNs (even in ResNet), as a network becomes deeper, gradients of processing layers tend to be smaller.
As a result, that learning does not progress well.
Since a shallow network is less affected by the vanishing gradient problem,
ResDrop makes the network apparently shallow in learning by introducing a regularizer called Stochastic Depth;
it treats some of residual blocks stochastically selected as the identity mapping which directly outputs the input. \\
PyramidNet is another ResNet improvement which was the previous state-of-the-art~\cite{arXiv:1610.02915}.
ResNet has a few special residual blocks where the number of channels greatly increases.
It has been reported that they can interfere with learning ability.
On PyramidNet, the number of channels increases step by step on each reasidual block instead of these special residual blocks, so that it improves the accuracy.
Though the authors of PyramidNet point out that use of stochastic regularizers such as Dropout~\cite{JMLR:v15:srivastava14a} and the stochastic depth could improve the performance, we could not confirm the effect on the use of the stochastic depth. \\
In this paper, we propose a method to combine the stochastic depth of ResDrop and PyramidNet successfully.

\section{Related work}

\subsection{ResNet}
It is reported that the accuracy of DCNNs is saturated by the vanishing gradient problem.
It is caused by ``too much learning" in the convolutional layers; it means learning lasts regardless of no hope to improve the accuracy.
Since the ``too much learinng" is caused by the lack of ability to realize the identity mapping given as
\begin{equation}
\label{eq:eq1}
H(x) = x,
\end{equation}
where $x$ and $H(x)$ are the input and the output of a convolutional layer, respectively.
However, it is hard on convolutional layers to learn the identity mapping.
In order to handle the identity mapping more easily, the residual block has been proposed.
It is expressed as
\begin{equation}
\label{eq:eq2}
H(x) = F(x) + x,
\end{equation}
where $H(x)$ is the function of the residual block to the input $ x $ and $F(x)$ is the residual function to be fit by a few stacked layers of the residual block.
Eq.~\eqref{eq:eq2} can be learned more easily than Eq.~\eqref{eq:eq1}.
In fact, ResNet consisting of residual blocks has improved accuracy considerably compared with the conventional DCNNs.

\subsection{ResDrop}
ResNet cannot avoid the vanishing gradient problem when it becomes deeper because residual blocks do not realize the identity mapping perfectly.
Hence, a regularizer, stochastic depth, is introduced to ResNet; it probabilistically makes $F(x)=0$ for dealing with some residual blocks as the identity mapping.
It has been confirmed that the accuracy is improved by introducing the stochastic depth to ResNet.

\subsection{PyramidNet}
PyramidNet is another ResNet improvement which was the previous state-of-the-art~\cite{arXiv:1610.02915}.
ResNet has a few special residual blocks where the number of channels greatly increases.
It has been reported that they can interfere with learning ability.
On PyramidNet, the number of channels increases step by step on each reasidual block instead of these special residual blocks, so that it improves the accuracy.

\begin{figure}[t]
	\begin{center}
		\includegraphics[height=4cm,width=5cm]{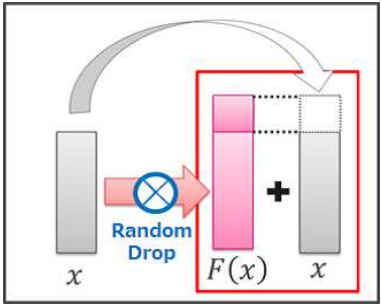}
		\caption{Residual block of PyramidDrop}
		\label{fig:PD}
	\end{center}
\end{figure}
\begin{figure}[t]
	\begin{center}
		\includegraphics[height=4cm,width=5cm]{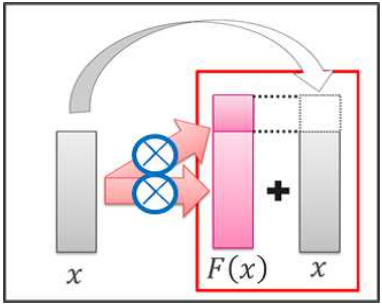}
		\caption{Residual block of PyramidSepDrop}
		\label{fig:PSD}
	\end{center}
\end{figure}

\section{Proposed Method}
Before presenting the proposed method, so as to highlight the advantage of the proposed method, we introduce a simple combination of the stochastic depth and PyramidNet; we call it Deep Pyramidal Residual Networks with Stochastic Depth (PyramidDrop).
PyramidDrop has residual blocks shown in Fig.~\ref{fig:PD}, where the cross in circle represents the random drop mechanism of the stochastic depth and $F(x)$ is the residual function with the mechanism.
As shown in Table~\ref{tab:Ex1},  PyramidDrop could not gain the accuracy as expected on a preliminary experiment.
Therefore, we propose Deep Pyramidal Residual Networks with Separated Stochastic Depth (PyramidSepDrop).
PyramidSepDrop has residual blocks shown in Fig.~\ref{fig:PSD}, where the $F(x)$ is separated into the upper part for increased channels and the lower part which has the same channels as the input.
The random drop mechanism of the stochastic depth is used in both parts.

\begin{table}[t]
  \begin{center}
    \begin{tabular}{|c||r|r|} \hline
      Method & CIFAR-10 & CIFAR-100 \\ \hline \hline
      PyramidNet~\cite{arXiv:1610.02915} & 3.77\% & 18.29\% \\
      PyramidDrop & 3.99\% & 18.30\% \\
      PyramidSepDrop & \bf 3.66\% & \bf 18.01\% \\ \hline
    \end{tabular}
    \caption{Error rates of PyramidNet, PyramidDrop and PramidSepDrop.}
    \label{tab:Ex1}
  \end{center}
\end{table}

\begin{figure}[t]
	\begin{center}
		\includegraphics[width=1.0\hsize]{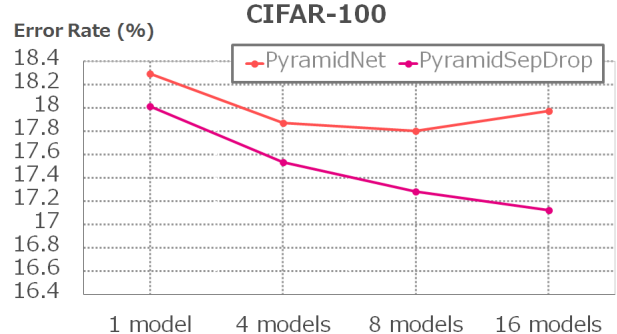}
		\caption{Error rates of PyramidSepDrop on the last epoch when increasing the number of models used for multi-model learning.}
		\label{fig:Ex2}
	\end{center}
\end{figure}

\section{Experiments}
\subsection{Conditions}
Regarding the preprocessing of images and learning conditions, we followed the experiments of PyramidNet~\cite{arXiv:1610.02915};
Networks were trained using the backpropagation by Stochastic Gradient Descent (SGD) with Nesterov momentum for 300 epochs on the CIFAR-10 and CIFAR-100 datasets.
The initial learning rate was set to 0.5, decayed by a factor of 0.1 at 150 and 225 training epochs, respectively.
The filter parameters were initialized by MSRA~\cite{arXiv:1502.01852}.
We used a weight decay of 0.0001, dampening to 0, and momentum of 0.9, with a batch size of 128. \\
We set parameters $ \alpha $ and Death Rate according to PyramidNet and ResDrop.
We adjusted the parameter $ \alpha $ used in PyramidNet to be $ \alpha = 5 * (depth- 2) / 6 $, where $depth$ is the number of layers, in order to increase 5 channels per residual block.

\begin{table}[t]
  \begin{center}
    \begin{tabular}{|c|c|c||r|r|} \hline
      $model$ & $depth$ & $\alpha$ & CIFAR-10 & CIFAR-100 \\ \hline \hline
      4 & 110 & 90 & 3.61\% & 17.53\% \\
      4 & 146 & 120 & {\bf 3.41\%} & 17.02\% \\
      4 & 182 & 150 & {\bf 3.45\%} & {\bf 16.33\%} \\ \hline
    \end{tabular}
    \caption{Error rates of PyramidSepDrop on the last epoch when increasing the number of layers.}
    \label{tab:Ex3}
  \end{center}
\end{table}

\begin{table}[t]
  \begin{center}
    \begin{tabular}{|c|c|c||r|r|} \hline
      $model$ & $depth$ & $\alpha$ & CIFAR-10 & CIFAR-100 \\ \hline \hline
      4 & 182 & 150 & 3.45\% & 16.33\% \\
      16 & 182 & 150 & {\bf 3.31\%} & {\bf 16.18\%} \\ \hline
    \end{tabular}
    \caption{Error rates of PyramidSepDrop on the last epoch when increasing the number of models on $ depth = 182 $.}
    \label{tab:Ex4}
  \end{center}
\end{table}

\begin{table}[t]
  \begin{center}
    \begin{tabular}{|c|c||r|r|} \hline
      Method & $depth$ & CIFAR-10 & CIFAR-100 \\ \hline \hline
      ResNet~\cite{He_CVPR2016} & 110 & 6.43\% & 25.16\% \\
      ResDrop~\cite{arXiv:1603.09382} & 110 & 5.23\% & 24.58\% \\
      DenseNet~\cite{arXiv:1608.06993} & 100 & 3.74\% & 19.25\% \\
      PyramidNet~\cite{arXiv:1610.02915} & 110 & 3.77\% & 18.29\% \\
      ResNeXt~\cite{arXiv:1611.05431} & 29 & 3.58\% & 17.31\% \\
      PyramidSepDrop & 182 & \bf 3.31\% & \bf 16.18\% \\ \hline
    \end{tabular}
    \caption{Error rates of the conventional methods and PyramidSepDrop.}
    \label{tab:FamousNets}
  \end{center}
\end{table}

\subsection{Results}
We conducted following four experiments. \\
1) We compared the error rates of PyramidNet, PyramidDrop and PyramidSepDrop using the same parameters; $depth$ was 110 and $\alpha$ was $90$.
The error rates of PyramidNet, PyramidDrop, PyramidSepDrop are shown in Table~\ref{tab:Ex1}.
We found that PyramidSepDrop, the proposed method, did gain the accuracy, while PyramidDrop did not. \\
2) We compared the error rates of PyramidNet and PyramidSepDrop on multi-model learning.
On the experiment, the input was a mini-batch of $N$ samples, which were divided into $N / model$, where $model$ is the number of mdoels, sub-batches and sent to each model separately for training, and the network parameters were communicated across the models in the process.
The result when increasing the number of models used for multi-model learning is shown in Fig.~\ref{fig:Ex2}.
The figure shows that as the number of models increased, the difference between PyramidNet and the PyramidSepDrop became larger.
Regarding the experiment on the number of models, the error rate increased in 16 models of PyramidNet, but the error rate decreased in PyramidSepDrop. \\
3) Table~\ref{tab:Ex3} shows the result when increasing the number of layers in PyramidSepDrop under the condition that the number of models was 4.
As a result, the error rates tended to decrease as the number of layers increased. \\
4) In the PyramidSepDrop, when the number of models was increased, the result when the number of layers was 182 is shown in the Table~\ref{tab:Ex4}.
As a result, the error rate was the lowest on $model = 16 $.

\section{Conclution}
We propose a method PyramidSepDrop based on PyramidNet to which a stochastic networks similar to ResDrop is introduced.
As shown in the Table~\ref{tab:FamousNets}, PyramidSepDrop greatly reduced the error rate from the conventional methods.

\section{Acknowledgement}
This work is partly supported by JSPS KAKENHI \#25240028 and JST CREST.

\end{document}